\documentclass{article}
\usepackage{iclr2027_conference,times}

\usepackage{amsmath,amssymb}
\usepackage{graphicx}
\usepackage{subcaption}
\usepackage{booktabs}
\usepackage{placeins}
\usepackage{algorithm}
\usepackage{algpseudocode}
\usepackage{microtype}
\usepackage{xcolor}
\usepackage{hyperref}
\usepackage{xurl}

\iclrfinalcopy 

\title{MIPS: Mitigating Mode Collapse in\\ Reinforcement Learning for Large Outcome Spaces}
\author{
Armaan Sandhu$^{1}$\thanks{Equal contribution.} \quad
Abhijeet Sinha$^{2}$\footnotemark[1] \quad
Dianbo Liu$^{2}$ \\
\textnormal{$^{1}$University of Massachusetts Amherst \quad
$^{2}$National University of Singapore} \\
\texttt{apsandhu@umass.edu, \{abhijeet, dianbo\}@nus.edu.sg}
}

\newcommand{\draftfigure}[2]{%
  \IfFileExists{#1}{%
    \includegraphics[width=\textwidth]{#1}%
  }{%
    \fbox{\parbox[c][1.35in][c]{\dimexpr\linewidth-2\fboxsep-2\fboxrule\relax}{%
      \centering\textbf{Figure asset pending}\\[0.5\baselineskip]
      #2\\[0.5\baselineskip]
      \texttt{\detokenize{#1}}}}%
  }%
}

\begin{document}
\raggedbottom
\maketitle

\begin{abstract}
Expected-reward maximization in reinforcement learning concentrates probability mass on a few outcomes, even in tasks that require many distinct, high-quality solutions, a failure known as mode collapse.
Inverse Probability Scaling (IPS) counters this in group-based policy gradients such as GRPO by dividing each reward by the outcome's probability, estimated from its frequency within the sampled group.
When the outcome space is large relative to the group, outcomes rarely repeat and these estimates become uninformative.
Since an outcome can be constructed along many trajectories, its probability is a sum over all of them, and computing it directly is intractable.
We propose \textbf{Marginal IPS (MIPS)}, which performs marginal evaluation by combining the sampled trajectory with a learned model of the trajectories that lead to each outcome.
On Hypergrids, where the exact sampling distribution is computable, MIPS stays near the reward-distribution target as the outcome space grows relative to the rollout group, while frequency-based IPS degrades.
In molecular synthesis and phylogenetic inference, MIPS sustains diverse, high-quality sampling where GRPO and frequency-based IPS collapse, showing that MIPS serves as a reliable solution to mode collapse in RL.
\end{abstract}

\section{Introduction}

Many reinforcement learning (RL) problems require a policy to preserve
multiple distinct, high-quality solutions rather than identify a single
optimum. This setting arises in large-language-model post-training
\citep{ouyang2022training}, molecular discovery
\citep{zhu2023molecular,koziarski2024rgfn}, and combinatorial generation
\citep{bengio2021gfn}, where several terminal outcomes may be simultaneously
useful \citep{kirk2024diversity,pan2023gafn}. Standard expected-reward
optimization in RL, however, often concentrates probability mass on a small subset
of rewarding outcomes, yielding repetitive or homogeneous samples despite
high expected reward
\citep{sinha2026ips,kirk2024diversity,olivecrona2017molecular,setpo2026}. We
refer to this concentration of the terminal sampling distribution as
\emph{mode collapse}.

To mitigate mode collapse in RL, we seek a policy that maintains broad coverage of useful outcomes while assigning greater probability to higher-quality ones.
This requires the learned outcome probabilities to reflect the reward landscape, so that distinct rewarding regions remain represented according to their relative quality.
Under this view, diversity depends not only on how broadly probability mass is spread, but also on where it is placed in the outcome space.
This turns mode-collapse mitigation into learning reward-aligned outcome probabilities, i.e.,
\(p_\theta(o_i)\propto R(o_i)\), rather than optimizing generic diversity proxies such as policy entropy, novelty, or pairwise disagreement.
Generic entropy regularization can increase stochasticity without specifying how probability should be allocated across outcomes \citep{haarnoja2018sac}, although specialized formulations can induce reward-aligned behavior \citep{tiapkin2024entropygfn,deleu2024multipath}.
GFlowNets provide an established framework for learning such outcome probabilities \citep{bengio2021gfn,bengio2023foundations}; our focus is on achieving this behavior within a conventional RL optimization pipeline.

Inverse Probability Scaling (IPS) provides such a route by scaling each
sampled reward by the inverse probability of its terminal outcome
\citep{sinha2026ips}. Intuitively, IPS increases the relative advantage of
rare rewarding outcomes and decreases that of outcomes sampled more
frequently.

The difficulty with IPS is estimating the marginal probability of a terminal outcome.
IPS-GRPO uses outcome frequencies within a rollout group, which become
uninformative for outcomes whose probability is too small to yield repeated
observations within the group (Section~\ref{sec:ips-failure}).

A readily available trajectory likelihood is not a substitute for the needed terminal probability. 
In constructive environments, many distinct action sequences can lead to the same outcome, so one path's probability does not include the probability of the alternative paths producing that outcome.

\begin{figure}[!htbp]
    \centering
    \includegraphics[width=\linewidth]{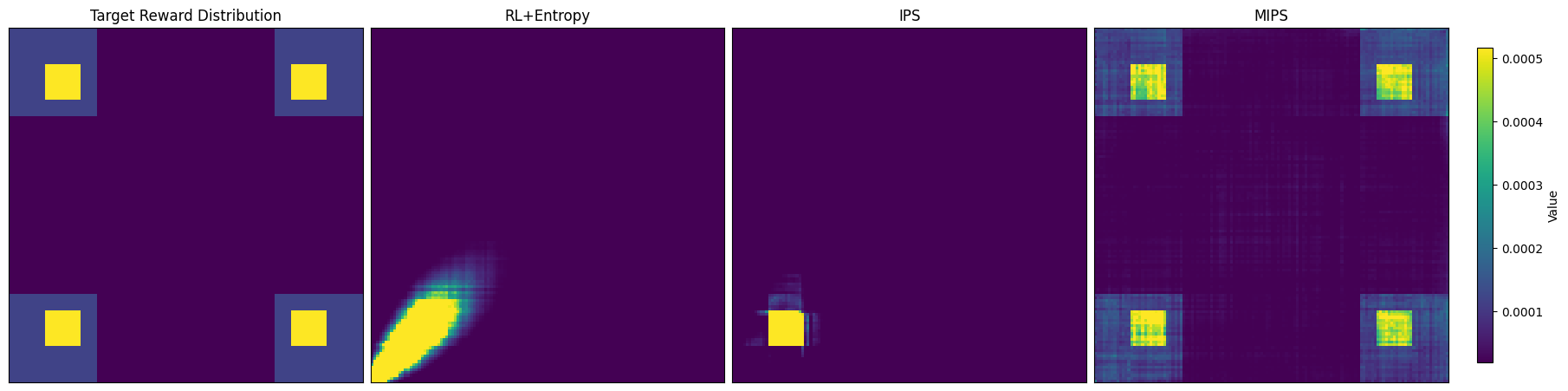}
    \caption{Terminal distributions on a Hypergrid. Left to right: the
    reward-distribution target, entropy-regularized RL, frequency-based
    IPS-GRPO, and MIPS-GRPO.}
    \label{fig:motivation}
\end{figure}

We introduce \emph{Marginal Inverse Probability Scaling} (MIPS), which evaluates the marginal terminal probability using one sampled trajectory together with a learned trajectory-conditional model. 
The conditional model accounts for the many-to-one mapping from trajectories to outcomes, allowing MIPS to estimate the outcome marginal without sampling a large number of trajectories or explicit path enumeration.
The resulting scaled rewards enter the standard GRPO surrogate unchanged \citep{shao2024deepseekmath,schulman2017ppo}. 
Figure~\ref{fig:motivation} illustrates the effect on a Hypergrid with four rewarding modes: entropy-regularized RL and frequency-based IPS each collapse onto a single mode, whereas MIPS places probability on all four in line with the reward.

We evaluate MIPS across three increasingly large and structured domains. 
On enumerable Hypergrids, exact policy probabilities allow direct measurement of agreement between the reward landscape and policy; sweeps over terminal-space size and rollout-group size additionally test how the effectiveness of IPS depends on group size $G$ as the outcome
space grows.
In reaction-based molecular synthesis \citep{koziarski2024rgfn}, we measure retention of diverse high-reward modes and molecular scaffolds. 
In phylogenetic tree construction \citep{zhou2024phylogfn}, we measure terminal coverage and alignment between sampling mass and reward. 
The latter two evaluations assess diverse coverage and the retention of high-quality solutions.

Our contributions are: 

\begin{itemize}
    \item We introduce MIPS, which estimates an outcome's marginal probability using a learned trajectory-conditional model, while retaining the standard GRPO update.
    \item We evaluate MIPS on Hypergrids, where the learned outcome distribution can be compared exactly with the reward distribution. We then extend the evaluation to molecular synthesis and phylogenetic tree construction, where we measure coverage of high-quality outcomes and alignment with the reward.
\end{itemize}

\section{Why IPS has limitations}
\label{sec:ips-failure}

Inverse Probability Scaling (IPS) modifies the reward for an outcome \(o\) as
\[
\widetilde{R}(o)=\frac{R(o)}{P_\theta(o)},
\]
where \(P_\theta(o)\) is the probability that the current policy generates \(o\). With the true \(P_\theta(o)\), IPS admits a reward-proportional stationary solution,
\(P_\theta(o_i)\propto R(o_i)\) \citep{sinha2026ips}.

In practice, however, \(P_\theta(o)\) is generally unknown. IPS-GRPO estimates it from outcome frequencies within a group of \(G\) rollouts \citep{sinha2026ips}, so the estimate has resolution \(1/G\). For an outcome \(o\), the expected number of occurrences in a group is \(G P_\theta(o)\), and after one occurrence, the probability of observing it again is $1-\left(1-P_\theta(o)\right)^{G-1}$.
When \(G P_\theta(o)\ll 1\), repeated observations are unlikely, and a sampled outcome is therefore typically observed only once and assigned probability \(1/G\). In this regime, the within-group frequency provides little information about its underlying probability. 
This makes the choice of \(G\) important in practice.

In large constructive problems, the size of the outcome space is typically unknown in advance, and the effective outcome space accessible to the policy evolves throughout training. Because neither the scale nor the effective support of this space is fixed, no single choice of \(G\) can reliably provide enough repeated outcomes throughout training. Choosing \(G\) to be conservatively large is also impractical because the sampling cost grows directly with the group size, motivating an estimator that does not depend on repeated occurrences of the same outcome.

MIPS addresses this limitation by estimating the marginal probability of each sampled outcome without relying on its frequency within the rollout group.

\section{Marginal Inverse Probability Scaling (MIPS)}

IPS requires the probability of the outcome $o$, rather than the probability
of any particular trajectory that generates it. For an outcome $o$, this
\emph{marginal outcome probability} is
\[
P_\theta(o)
=
\sum_{\tau\in\mathcal T(o)} P_\theta(\tau),
\]
where $\mathcal T(o)$ denotes the set of construction trajectories generating
output $o$. This marginalization motivates the name \emph{Marginal Inverse
Probability Scaling} (MIPS). A single trajectory likelihood cannot generally
replace this quantity because multiple distinct trajectories may generate the same outcome. In large constructive spaces, explicitly summing over all
trajectories in $\mathcal T(o)$ is generally intractable.

For a finite, acyclic, deterministic constructive process with positive rewards, \(R(o)>0\), MIPS avoids this sum using a learned trajectory-conditional model.

Let $\tau=(s_0,a_0,\ldots,s_T)$ be an on-policy trajectory with likelihood $P_\theta(\tau)=\prod_{t=0}^{T-1}\pi_\theta(a_t\mid s_t)$.
Since $o$ is determined by $\tau$, $P_\theta(\tau,o)=P_\theta(\tau)$.
By the chain rule, $P_\theta(\tau,o)=P_\theta(o)P_\theta(\tau\mid o)$,
and therefore $P_\theta(\tau)=P_\theta(o)P_\theta(\tau\mid o)$.
For any supported trajectory,
\[
P_\theta(o)
=
\frac{P_\theta(\tau)}
     {P_\theta(\tau\mid o)}.
\]

Thus, given the trajectory-conditional probability, the marginal outcome probability can be recovered using one trajectory for the sampled outcome.

MIPS therefore introduces a learnable \textbf{trajectory-conditional model $c_\phi(\tau\mid o)$}, such that:
\[
c_\phi(\tau\mid o)
\approx
P_\theta(\tau\mid o),
\qquad and \qquad
\sum_{\tau\in\mathcal T(o)}
c_\phi(\tau\mid o)=1,
\]
and substitutes it into the marginal identity. For each sampled trajectory,
MIPS defines
\begin{equation}
\label{eq:mips-marginal-estimator}
P_{\theta,\phi}(o)=\frac{P_\theta(\tau)}{c_\phi(\tau\mid o)},
\qquad
\widetilde R=\frac{R(o)}{ P_{\theta,\phi}(o)}=
\frac{R(o)c_\phi(\tau\mid o)}{P_\theta(\tau)}.
\end{equation}

If $c_\phi(\tau\mid o)=P_\theta(\tau\mid o)$, then
$ P_{\theta,\phi}(o)=P_\theta(o)$ for every trajectory generating $o$.

To make the conditional model tractable, it can be factorized over predecessor
transitions. For a finite acyclic construction process with a unique initial
state and a Markov policy,
\[
c_\phi(\tau\mid o)
=
\prod_{t=1}^{T}
c_\phi(s_{t-1},a_{t-1}\mid s_t),
\]
with each factor normalized over valid incoming transitions.

The conditional model is trained by maximum likelihood using on-policy
trajectories:
\begin{equation}
\label{eq:mips-cond-loss}
\mathcal L_{\mathrm{cond}}(\phi)=-\mathbb E_{\tau\sim\pi_\theta}
\left[
\sum_{t=1}^{T}
\log c_\phi(s_{t-1},a_{t-1}\mid s_t)
\right].
\end{equation}
At each training step, MIPS uses the current conditional model $c_\phi(\tau\mid o)$ to compute scaled rewards for the new rollouts. 
It then keeps those scaled rewards fixed, normalizes them within the rollout group, and updates the policy with the usual clipped GRPO objective. The complete update is given in
Appendix~\ref{app:grpo-update}.

With an approximate conditional model, $ P_{\theta,\phi}(o)$ may vary across trajectories to the same outcome and is not generally an unbiased estimate of $P_\theta(o)$; nor do the samplewise estimates necessarily form a normalized outcome distribution.
Its accuracy depends on conditional-model fit and adequate policy support.

\subsection{Relation to GFlowNets}
\label{sec:relation-gfn}
GFlowNets also target reward-proportional terminal sampling \citep{bengio2021gfn,bengio2023foundations}. 
The trajectory-conditional factorization used by MIPS is similar to the backward policy used in GFlowNets \citep{malkin2022tb}. 
Related work connects GFlowNet objectives to entropy-regularized RL \citep{tiapkin2024entropygfn}, analyzes multi-path marginal corrections \citep{deleu2024multipath}, and develops policy-gradient and PPO-style GFlowNet training \citep{niu2024gfnpg,zykovamyzina2026ppo}.

MIPS differs in how this conditional distribution is used: it estimates the
\emph{current policy's marginal terminal probability} for IPS, thereby
avoiding repeated terminal observations while retaining the standard GRPO
surrogate. It does not introduce a new GFlowNet objective. Detailed
comparisons appear in Appendix~\ref{app:related}.

\section{Experiments}
\label{sec:Experiments}
We first study MIPS in a controlled setting, the Hypergrid environment.
Because the terminal distribution of each trained policy can be computed exactly in this setup, we can compare it directly with the reward-distribution target and vary the outcome-space size and group size independently.
We then test whether these findings carry over to two large structured domains, molecular synthesis and phylogenetic tree construction, where the outcome spaces are too large to enumerate.
In these domains, we assess coverage of high-quality outcomes and the alignment between sampling density and reward, and include task-specific GFlowNet samplers as reference points.

\subsection{Hypergrid scalability with outcome-space size and group size}
\label{sec:hypergrid-scaling}

We use a two-dimensional Hypergrid with side length $H$, in which each state is represented by a pair of coordinates in $\{0,\ldots,H-1\}^2$. 
An agent starts at $(0,0)$ and, at each step, the agent can either increment one coordinate or stop. 
Any state at which the agent stops is a terminal state, so the agent can choose to terminate at any point in the grid. 
Different orders of coordinate increments can reach the same terminal state, creating multiple construction trajectories for each outcome. 
A grid with side length $H$ has $H^2$ possible terminal states. Here the outcomes are the same as the terminal states.
Since the Hypergrid is enumerable, we can compute the learned outcome distribution $P_\theta$ exactly and directly assess whether it matches the desired reward-aligned distribution.
Each episode terminates at a grid coordinate \(s\in\{0,\ldots,H-1\}^2\), which is mapped uniformly to \(x\in[-1,1]^2\).
The reward is defined as
\begin{equation}
R(x)
=R_0
+R_1\prod_{i=1}^{2}{1}\!\left(|x_i|>0.5\right)
+R_2\prod_{i=1}^{2}{1}\!\left(0.6<|x_i|<0.8\right),
\end{equation}
where $0<R_0\ll R_1<R_2$.
This construction produces $4$ symmetric high-reward modes surrounded by a ring, resulting in a sharply multimodal reward landscape (Figure~\ref{fig:motivation}, left). We set $R_0=0.1$, $R_1=0.5$, and $R_2=2.0$, ensuring that every terminal state receives a positive reward.

\textbf{Experiment.}
We compare three training methods: GRPO, IPS-GRPO, and MIPS-GRPO. 
For each method, we consider grid side lengths $H\in\{4,8,16,32,64,128,256,512\}$ and rollout-group sizes $G\in\{16,32,64,128,256,512,1024,2048\}$, yielding $192$ experimental configurations in total. 
For each configuration, we trained the policy until convergence.
To quantify how closely the trained policy's outcome distribution matches the reward distribution, we use Jensen--Shannon divergence (JSD) and total variation distance (TV), with lower values indicating better alignment.
Because the environment is enumerable, we use dynamic programming to compute the exact outcome distribution of each trained policy. 

\begin{figure}[!htbp]
    \centering
    \includegraphics[width=\linewidth]{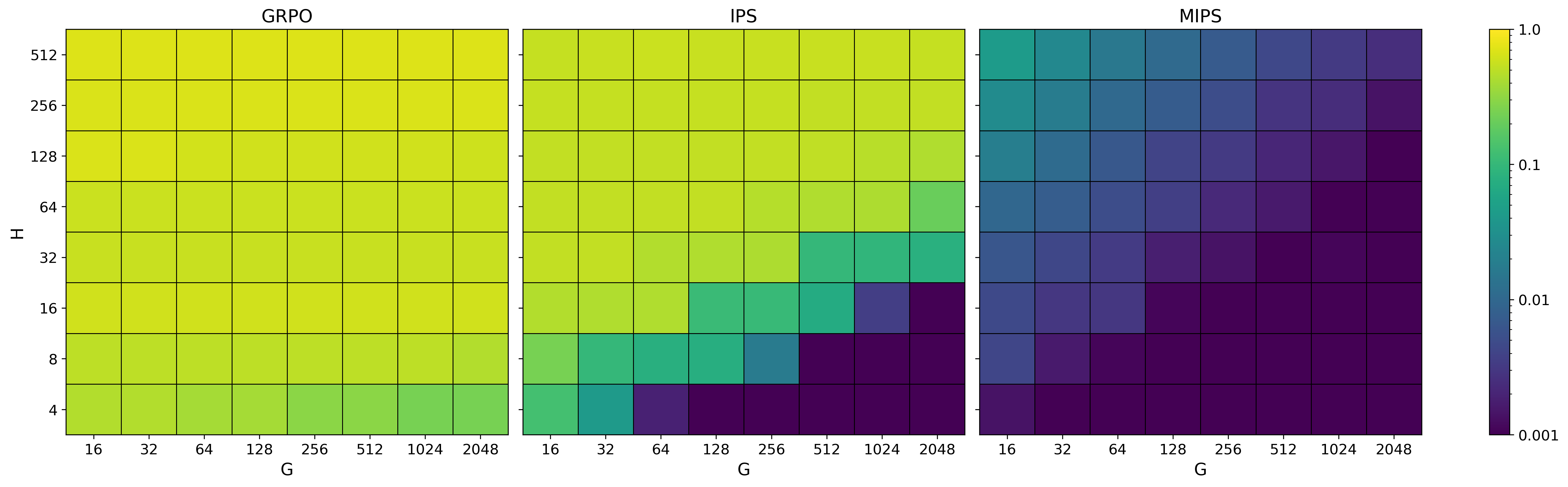}
    \caption{\textbf{Hypergrid scalability.} JSD between the exact outcome distribution and the reward-distribution target
    for GRPO (left), IPS-GRPO (center), and MIPS-GRPO (right). Grid side length $H$ is shown on the vertical
    axis and group size $G$ on the horizontal axis, with $H^2$ outcomes and a shared logarithmic color scale across panels.}
    \label{fig:hypergrid-scalability}
\end{figure}

\paragraph{Results.}
Figure~\ref{fig:hypergrid-scalability} shows how the three methods behave as the outcome space becomes larger and the rollout group size changes.
GRPO remains far from the reward-distribution target across most settings.

IPS behaves differently. 
On smaller outcome spaces, it can closely match the target reward distribution, especially when the rollout group is sufficiently large. 
However, its performance deteriorates as the outcome space grows.
Increasing the group size helps, but mainly shifts the point at which this deterioration occurs rather than removing it. 
MIPS shows a different scaling behavior. 
Its outcome distribution remains close to the reward-distribution target over a much wider range of outcome spaces and group sizes. 
In particular, its behavior changes much less when the group is small relative to the number of possible outcomes.

\begin{figure}[!ht]
    \centering
    \begin{subfigure}{0.8\linewidth}
        \centering
        \includegraphics[width=\linewidth]{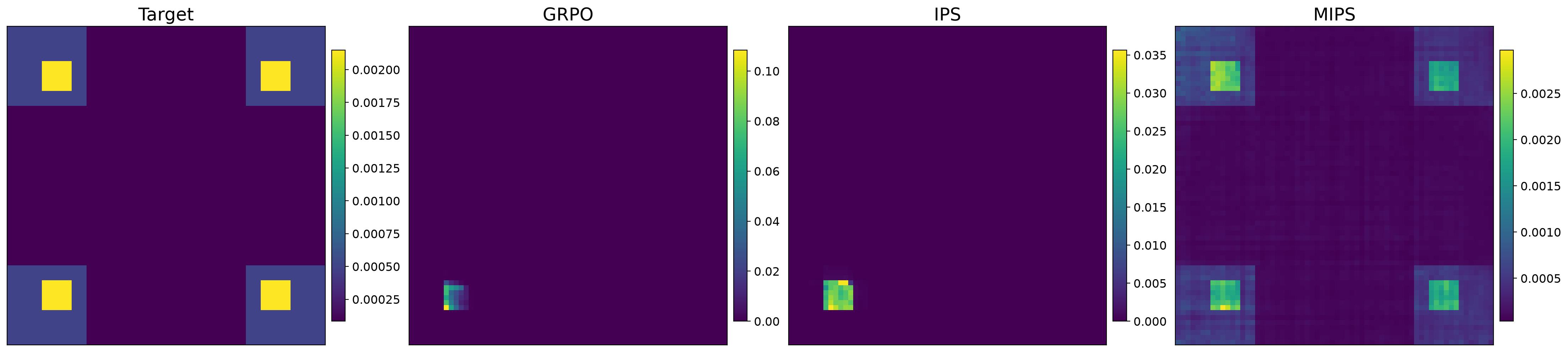}
        \caption{$G=64$}
    \end{subfigure}
    \vspace{-0.1\baselineskip}

    \begin{subfigure}{0.8\linewidth}
        \centering
        \includegraphics[width=\linewidth]{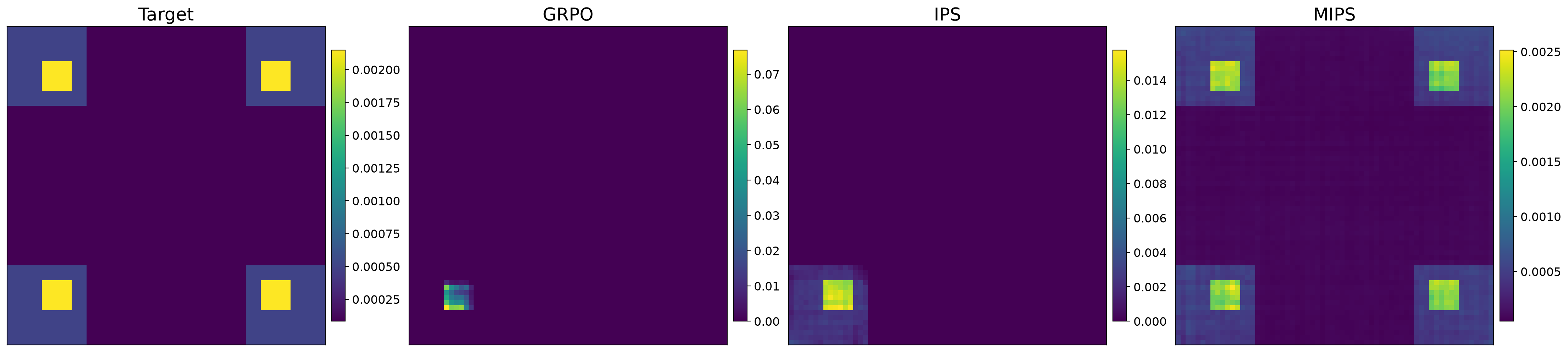}
        \caption{$G=512$}
    \end{subfigure}
    \vspace{-0.1\baselineskip}

    \begin{subfigure}{0.8\linewidth}
        \centering
        \includegraphics[width=\linewidth]{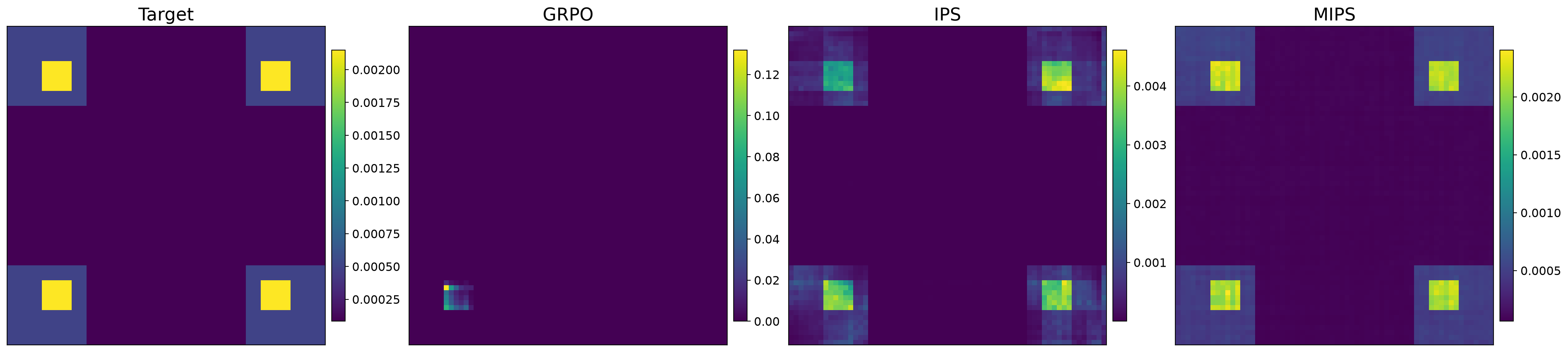}
        \caption{$G=2048$}
    \end{subfigure}
    \caption{\textbf{Outcome distributions on the $64\times64$ Hypergrid.}
    The reward-distribution target and the exact outcome distributions learned
    by GRPO, IPS-GRPO, and MIPS-GRPO are shown for increasing
    rollout-group sizes $G$.}
    \label{fig:hypergrid-distributions}
\end{figure}

Figure~\ref{fig:hypergrid-distributions} makes this behavior more concrete by showing the learned outcome distributions for a fixed Hypergrid while the group size is increased. 
GRPO concentrates almost all of its probability around a single rewarding region, exhibiting mode collapse. 
With a small group, IPS-GRPO shows a similar form of collapse but covers more outcomes.
As the group size $G$ becomes larger, IPS becomes more effective. 
MIPS behaves differently. Even with the smaller group size $G$, it places probability around all rewarding regions. 
Increasing the group size $G$ mainly refines this distribution rather than changing which modes are represented.

The important observation is therefore not simply that MIPS produces a more diverse distribution, but that its diversity follows the structure of the reward distribution itself.
The complete JSD and TV results (Appendix~\ref{app:hypergrid-sweep}, Tables~\ref{tab:hypergrid-jsd} and~\ref{tab:hypergrid-tv}) show that, although sufficiently large groups allow IPS-GRPO to match MIPS on the smallest grids, MIPS achieves consistently better alignment as the outcome space grows, across all tested group sizes $G$.

Taken together, the Hypergrid experiments verify two key properties of MIPS.
First, MIPS preserves diversity as the outcome space grows without requiring the rollout group size $G$ to grow with it. 
Second, this diversity is not produced by indiscriminate randomness: the learned policy distribution remains structured by the reward itself, placing probability mass across the distinct rewarding modes rather than spreading it uniformly. 
These results show that MIPS can scale to large outcome spaces while maintaining reward-aligned diversity and substantially reducing sensitivity to the hyperparameter $G$.

\begin{figure}[!htbp]
    \centering
    \begin{subfigure}[t]{0.4\linewidth}
        \centering
        \includegraphics[width=\linewidth]{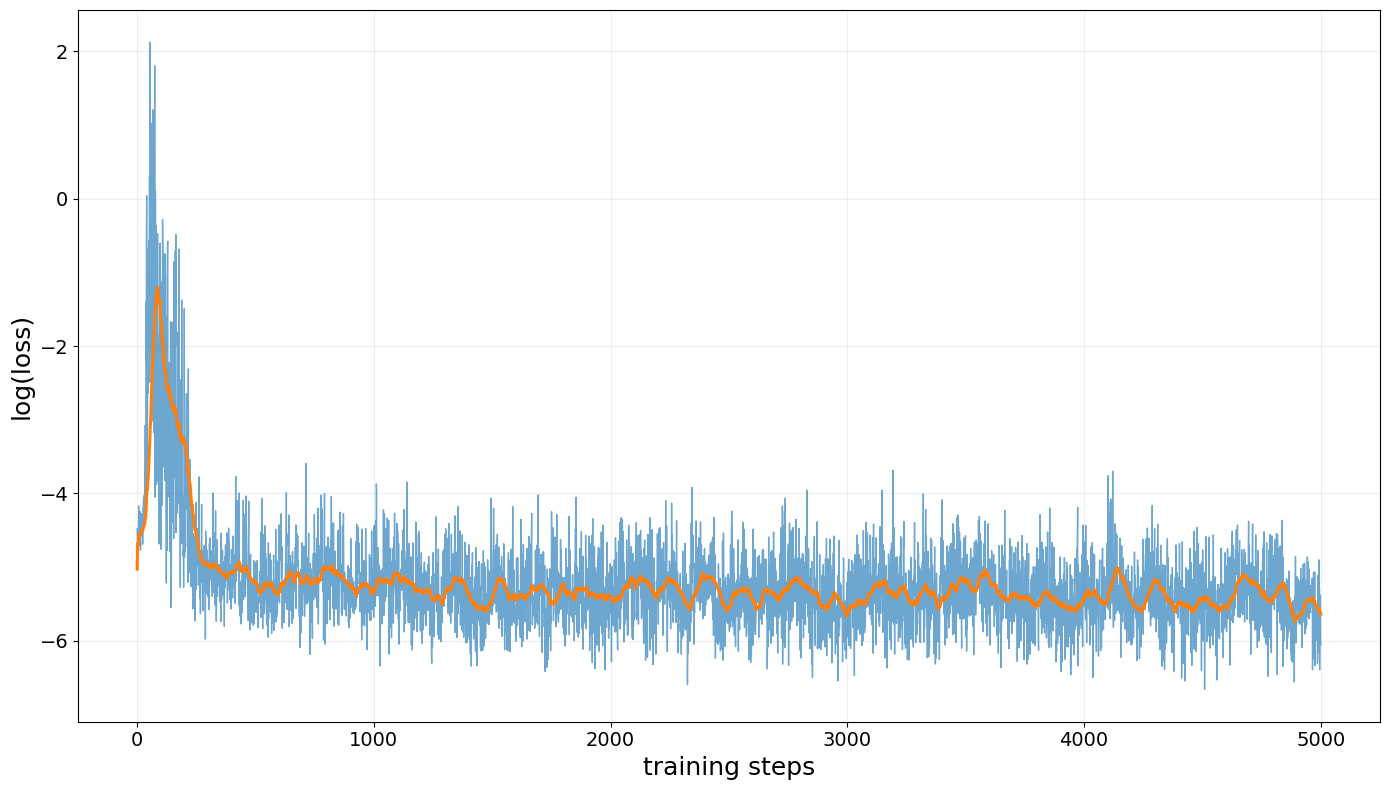}
        \caption{Log mean squared probability error.}
    \end{subfigure}
    \hspace{0.02\linewidth}
    \begin{subfigure}[t]{0.4\linewidth}
        \centering
        \includegraphics[width=\linewidth]{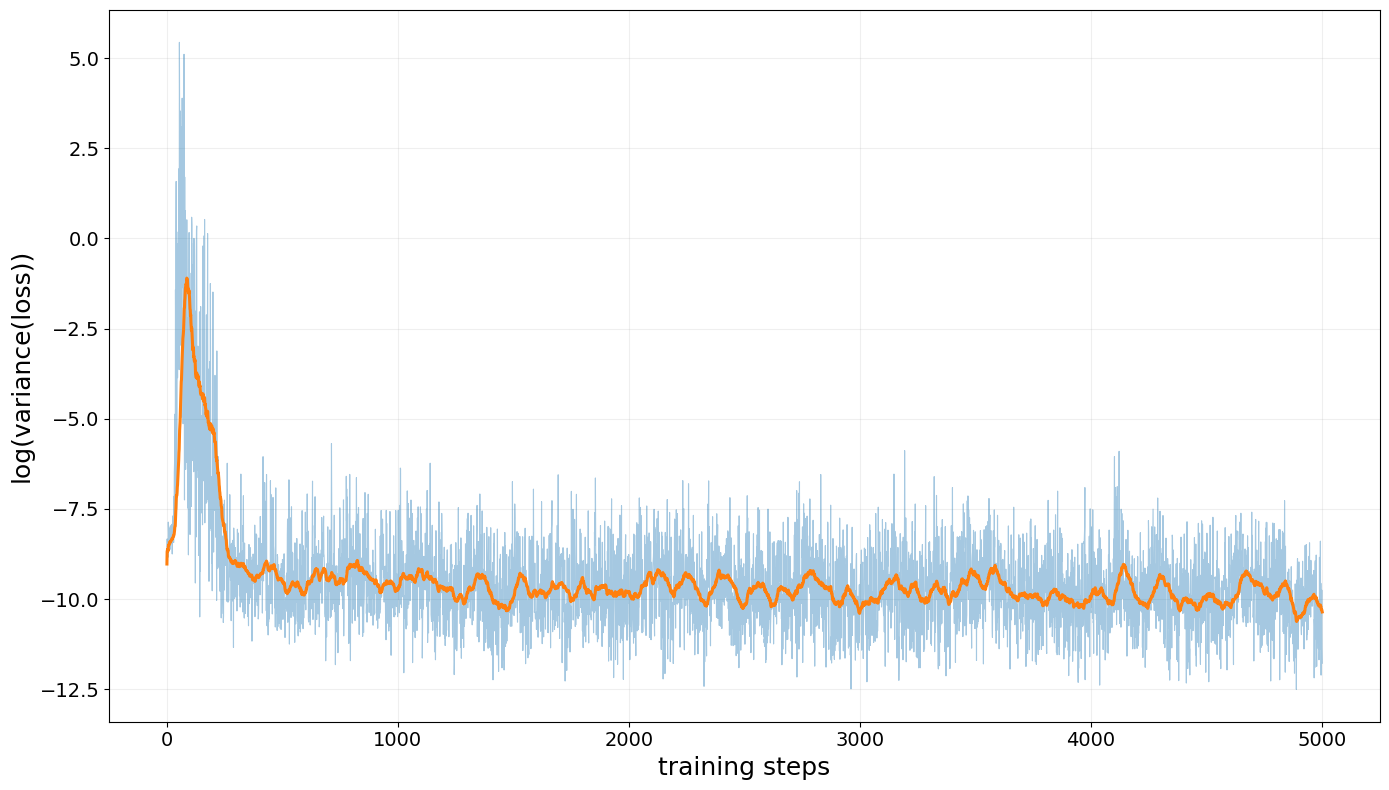}
        \caption{Log variance of the squared probability error.}
    \end{subfigure}

    \caption{\textbf{Marginal-probability estimation.} Raw measurements and smoothed trends are shown.}
    \label{fig:marginal-estimation}
\end{figure}

\paragraph{Stability and convergence of marginal-probability estimation.}
Because Hypergrid marginals are exactly computable by dynamic programming, we
directly evaluate the accuracy of the MIPS estimator during training. For each trajectory sampled during training, we measure
$
    \ell_i =
        \left(P_{\theta,\phi}(o_i)-P_\theta(o_i)\right)^2,
$
where $P_{\theta,\phi}(o_i)$ is the estimated marginal and $P_\theta(o_i)$ is
the exact marginal under the current policy. Figure~\ref{fig:marginal-estimation}
reports the batch mean and variance of $\ell_i$ across training on the log scale.

Both decrease rapidly after an initial transient and remain low thereafter.
This empirically supports stable convergence of the coupled policy and trajectory-conditional model to a low-error regime.

\subsubsection{Ablation of conditional scaling}
\label{sec:conditional-ablation}

We ablate the use of the trajectory-conditional model $c_\phi(\tau \mid o)$ in the MIPS scaling
factor. In the full method, the scaling term uses the conditional trajectory
model $c_\phi(\tau \mid o)$ to account for how trajectories are distributed
given the outcome. In the ablated variant, this conditional information is
removed from the scaling computation, so the estimator relies only on the
trajectory likelihood $P_\theta(\tau)$.
This is equivalent to setting \(c_\phi(\tau \mid o)=1\). Any other constant yields the same behavior because the scaled rewards are group-normalized when computing advantages.

Table~\ref{tab:conditional-ablation} shows that removing conditional scaling
substantially degrades alignment with the reward-distribution target across
all Hypergrid configurations. Both Jensen--Shannon divergence (JSD) and total
variation (TV) distance increase sharply in every setting.
These results show that the benefit of MIPS does not arise from trajectory probability scaling alone. 
Conditioning the scaling factor on the outcome is important for estimating the marginal outcome probability and, consequently, for matching the target reward distribution.

\begin{table}[htbp]
\centering
\small
\caption{\textbf{Ablation of conditional scaling on Hypergrids.} JSD and TV distance from the reward-distribution target. We compare full MIPS with a variant that does not use the trajectory-conditional model in the scaling factor.}
\label{tab:conditional-ablation}

\begin{tabular}{ccrrrr}
\toprule
& & \multicolumn{2}{c}{JSD $\downarrow$}
& \multicolumn{2}{c}{TV $\downarrow$} \\
\cmidrule(lr){3-4} \cmidrule(lr){5-6}
$H$ & $G$
& MIPS & No $c_\phi(\tau \mid o)$
& MIPS & No $c_\phi(\tau \mid o)$ \\
\midrule
32 & 32 & \textbf{0.0032} & 0.5740 & \textbf{0.0617} & 0.9147 \\
32 & 64 & \textbf{0.0040} & 0.5592 & \textbf{0.0697} & 0.9013 \\
64 & 32 & \textbf{0.0077} & 0.6315 & \textbf{0.0996} & 0.9622 \\
64 & 64 & \textbf{0.0052} & 0.6428 & \textbf{0.0815} & 0.9689 \\
\bottomrule
\end{tabular}
\end{table}

\subsection{Molecular synthesis}
\label{sec:molecules}

Molecular synthesis can be viewed as a sequential decision-making problem where an agent builds a molecule through a series of chemically valid reactions. 
At each step, the agent selects a reaction and compatible building blocks, and the final molecule receives a reward based on its desired properties.
This gives a natural reinforcement-learning formulation, where the sequence of synthesis decisions forms a trajectory and the resulting molecule is the outcome. 
The number of possible trajectories grows combinatorially with the available reactions, building blocks, and synthesis steps, creating an enormous search space. 
Moreover, different synthesis trajectories can lead to the same molecule, so many trajectories may correspond to a single outcome. 
The goal is therefore not only to find the best molecule, but to explore this large combinatorial space and discover diverse, high-quality molecules.

\paragraph{Environment.}
We use the reaction-based environment, sEH proxy task, and reward construction of RGFN \citep{koziarski2024rgfn} without modification.
Starting from an initial building block, the policy repeatedly selects a reaction template, a compatible building block, and one of the resulting products until choosing to stop. 
The chemistry contains $17$ reaction types and $350$ low-cost building blocks. 

Thus, as the number of reaction steps increases, many different trajectories can contribute probability to the same molecule, making this a challenging setting for marginal outcome-probability estimation. 
We compare GRPO, IPS-GRPO, MIPS-GRPO, and RGFN trained with its published trajectory-balance configuration.

\paragraph{Evaluation.}
After $4{,}000$ training iterations, we draw $50{,}000$ molecules from each final policy.
Molecules are identified by canonical SMILES, and uniqueness is measured only over valid samples. 
To measure diversity specifically within the high-quality region, we follow the mode criterion of \citet{koziarski2024rgfn}: unique molecules with sEH proxy score above $7$ are processed in decreasing score order, and the Leader algorithm selects a new mode when its Tanimoto similarity to every previously selected mode is at most $0.5$. 
We additionally count distinct Murcko scaffolds among molecules with proxy score above $7$ to measure whether high-quality samples span different molecular cores. 
Finally, we report the mean sEH proxy over all samples and over the discovered modes to assess whether broader coverage is accompanied by maintaining the molecular quality.

\begin{table}[htbp]
\centering
\small
\caption{\textbf{Final-policy molecular synthesis results.}
Statistics from $50{,}000$ samples across 3 seeds.
The selected metrics measure overall coverage, high-quality diversity,
structural diversity, and molecular quality.}
\label{tab:molecule-main}

\resizebox{\textwidth}{!}{%
\begin{tabular}{llcccc}
\toprule
Metric & What it measures & GRPO & IPS-GRPO & RGFN & MIPS-GRPO \\
\midrule

Unique molecules (count) $\uparrow$
& Overall molecular diversity
& $1 \pm 0$
& $17 \pm 11$
& $34{,}744 \pm 1{,}270$
& $\mathbf{45{,}794 \pm 2{,}332}$ \\

High-reward modes $\uparrow$
& Distinct high-quality regions
& $0 \pm 0$
& $0 \pm 0$
& $18 \pm 1$
& $\mathbf{70 \pm 2}$ \\

Murcko scaffolds (proxy $>7$) $\uparrow$
& High-quality structural diversity
& $0 \pm 0$
& $0 \pm 0$
& $25 \pm 6$
& $\mathbf{149 \pm 42}$ \\

Mean proxy (all samples)
& Overall molecular quality
& $6.54 \pm 0.37$
& $4.68 \pm 1.31$
& $4.19 \pm 0.37$
& $4.47 \pm 0.13$ \\

Mean proxy (modes) $\uparrow$
& Quality of discovered modes
& ---
& ---
& $7.17 \pm 0.05$
& $\mathbf{7.22 \pm 0.08}$ \\

\bottomrule
\end{tabular}%
}
\end{table}

\paragraph{Coverage and quality.}

Table~\ref{tab:molecule-main} compares the methods in terms of overall molecular diversity, high-reward mode coverage, structural diversity, and proxy quality.

GRPO collapses almost completely, producing essentially a single unique molecule per run. IPS-GRPO improves overall uniqueness modestly, generating $17$ unique molecules on average, but still fails to recover any high-reward modes. RGFN maintains much broader support, generating $34{,}744$ unique molecules ($69.5\%$ of valid samples) and recovering $18$ distinct high-reward modes.

MIPS-GRPO further expands this coverage, generating $45{,}794$ unique molecules ($91.8\%$ of valid samples), along with $70$ high-reward modes and $149$ high-reward Murcko scaffolds. The increase in scaffolds indicates that these additional modes span different molecular cores rather than consisting only of small variations of similar structures.

This broader coverage does not come at the cost of proxy quality. MIPS-GRPO achieves a mean proxy score of $4.47$ across all generated molecules, compared with $4.19$ for RGFN, while its discovered modes have a mean proxy score of $7.22$, compared with $7.17$ for RGFN. Together, these results show that MIPS-GRPO expands both overall and high-reward coverage while maintaining comparable molecular quality.

\subsection{Phylogenetic tree construction}
\label{sec:phylo}

Phylogenetic tree construction can be formulated as a sequential decision-making problem in which an agent builds a tree by repeatedly merging pairs of subtrees and assigning branch lengths. 
The resulting sequence of construction decisions forms a trajectory, while the completed phylogenetic tree is the outcome and receives a reward based on how well it explains the observed sequence data. 
This is a challenging combinatorial search problem because the number of possible tree topologies grows rapidly with the number of taxa, and many different construction trajectories can lead to the same final tree. 
The task therefore provides a structured setting in which to test whether a policy can retain diverse tree hypotheses while continuing to favor high-quality ones.

\paragraph{Setup and evaluation.}
We use a $10$-taxon setting derived from the DS1 benchmark and the bottom-up construction process of PhyloGFN \citep{zhou2024phylogfn}: starting from individual taxa, the policy merges pairs of subtrees and assigns branch lengths until a complete tree is formed. 
Rewards are based on the Jukes--Cantor log-likelihood \citep{jukes1969evolution}. 
We compare GRPO, IPS-GRPO, MIPS-GRPO, and PhyloGFN using $10^6$ samples from each final policy.

We measure coverage by the number of distinct tree signatures, where a signature combines tree topology and branch-length assignment, and assess tree quality using the mean log-likelihood (MLL) of sampled trees. Alignment with the reward distribution is measured by the Pearson correlation between outcome log-probability of the policy and log-likelihood of the generated tree, both on each method's own samples (sampling $r$), and on a common set of trees when the corresponding densities are available (shared-tree $r$).
For MIPS-GRPO, the tree log-probability is estimated using
Equation~\eqref{eq:mips-marginal-estimator}, with the trained policy and
conditional model providing $P_\theta(\tau)$ and $c_\phi(\tau\mid o)$.

\begin{table}[htbp]
\centering
\small
\caption{\textbf{Phylogenetic tree construction with $10$ taxa.}
Results from $10^6$ samples per final policy (mean $\pm$ standard deviation over three seeds).}
\label{tab:phylo-main}
\resizebox{\textwidth}{!}{%
\begin{tabular}{lcccc}
\toprule
Method
& Unique signatures / 1M $\uparrow$
& MLL
& Sampling $r$ $\uparrow$
& Shared-tree $r$ $\uparrow$ \\
\midrule
GRPO
& $1\pm0$
& $-4318.4\pm0.0$
& ---
& --- \\
IPS-GRPO
& $263{,}532\pm24{,}466$
& $-4294.4\pm3.7$
& $0.140\pm0.010$
& --- \\
PhyloGFN
& $1{,}000{,}000\pm0$
& $-4366.5\pm2.2$
& $0.836\pm0.044$
& $0.936\pm0.012$ \\
MIPS-GRPO
& $1{,}000{,}000\pm0$
& ${-4372.9\pm0.9}$
& $\boldsymbol{0.967\pm0.003}$
& $\boldsymbol{0.962\pm0.010}$ \\
\bottomrule
\end{tabular}%
}
\end{table}

\paragraph{Coverage and quality.}
Table~\ref{tab:phylo-main} shows that MIPS-GRPO and PhyloGFN achieve broad coverage, whereas IPS-GRPO covers far fewer trees and GRPO collapses to a single tree.
Broader coverage comes with lower MLL: instead of concentrating on a few high-likelihood trees, the policy spreads probability across a wider range of trees, including lower-likelihood ones.

\begin{figure}[htbp]
    \centering
    \includegraphics[width=\linewidth]{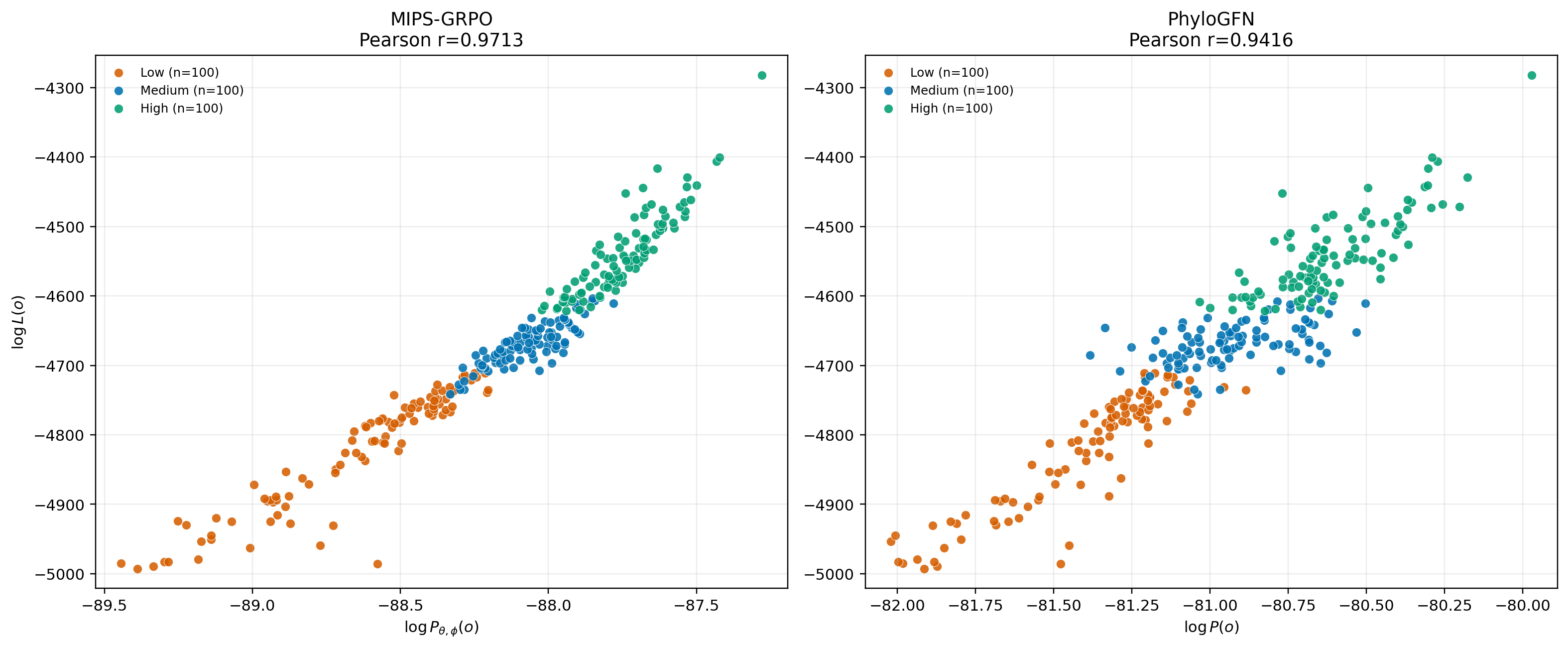}
    \caption{\textbf{Probability--likelihood alignment on a shared tree set.}
    Terminal log-likelihood versus outcome log-probability for MIPS-GRPO
    (left) and PhyloGFN (right) on a common set of $300$ trees ($50$ per
    method per log-likelihood band). A common set removes the dependence of
    the correlation on each method's own sampling distribution. MIPS-GRPO
    spreads its mass over a larger outcome space, so its log-probabilities
    are lower and absolute values are not comparable across panels. Pearson
    $r$ is reported above each panel.}
    \label{fig:phylo}
\end{figure}

MIPS-GRPO shows the strongest reward alignment, with a sampling correlation of $r=0.967$ compared with $0.836$ for PhyloGFN and $0.140$ for IPS-GRPO.
It also has the highest correlation on the shared tree set (Figure~\ref{fig:phylo}), so this advantage does not depend on each method's own samples.
Missing correlations could not be estimated because the corresponding policies generated too few distinct trees.
Overall, MIPS-GRPO combines broad topology coverage with a sampling density that increases consistently with tree likelihood.

Taken together, these results show that MIPS does more than increase stochasticity in phylogenetic construction.
Compared with GRPO and IPS-GRPO, it covers a far broader set of trees while assigning higher probability to higher-likelihood ones, and it achieves similarly broad coverage to PhyloGFN while exhibiting stronger reward alignment.

\FloatBarrier

\section{Discussion}
\label{sec:discussion}

Our results show that MIPS can maintain diverse, high-quality outcomes as the outcome space grows, effectively mitigating mode collapse in large outcome spaces. 
On Hypergrids, MIPS remains closely aligned with the reward distribution
across different outcome-space sizes and group sizes.
The marginal-estimation results and ablation studies further show that the trajectory-conditional model is important for accurately estimating marginal outcome probabilities.
This behavior also extends to larger, structured problems. 
In molecular synthesis, MIPS discovers diverse, high-reward molecules and molecular scaffolds while maintaining high mode quality. 
In phylogenetic inference, MIPS achieves broad tree coverage while assigning higher probability to higher-likelihood trees, resulting in stronger alignment between the policy and the reward distribution over outcomes. 

MIPS is limited by the accuracy of its learned trajectory-conditional model and by the assumptions underlying the current formulation, which considers finite, acyclic, deterministic constructive processes with positive rewards. 
Future work could extend MIPS beyond these assumptions, particularly to continuous outcome spaces in which outcomes may not be uniquely defined but instead exhibit varying degrees of similarity. 
Such settings could motivate marginal-probability corrections that account for similarity or local structure among outcomes rather than relying on exact outcome identity.

More broadly, this study opens the door to using RL algorithms effectively for search and sampling tasks, rather than only for reward maximization, through an appropriate choice of advantage calculation.

\bibliography{references}
\bibliographystyle{iclr2027_conference}

\clearpage
\appendix

\section{MIPS-GRPO policy update}
\label{app:grpo-update}

Algorithm~\ref{alg:mips-grpo} gives the update used for MIPS-GRPO.

\begin{algorithm}[htbp]
\caption{MIPS-GRPO}
\label{alg:mips-grpo}
\footnotesize
\begin{algorithmic}[1]
\Require policy $\pi_\theta$; conditional model $c_\phi$; group size $G$;
clipping parameter $\epsilon$; policy epochs $K$; conditional-model epochs $K_c$;
entropy coefficient $\beta$ for entropy $\mathcal H\!$
\For{each update}
    \State $\pi_{\mathrm{old}}\gets\pi_\theta$
    \State Sample $G$ trajectories
    $\tau_i=(s_{i,0},a_{i,0},\ldots,s_{i,T_i})\sim\pi_{\mathrm{old}}$
    \State Let $o_i$ denote the terminal outcome of $\tau_i$

    \Statex \textbf{without gradients:}
    \For{$i=1,\ldots,G$}
        \State $\ell_i^\pi \gets
        \sum_{t=0}^{T_i-1}
        \log\pi_{\mathrm{old}}(a_{i,t}\mid s_{i,t})$
        \State $\ell_i^c \gets
        \sum_{t=1}^{T_i}
        \log c_\phi(s_{i,t-1},a_{i,t-1}\mid s_{i,t})$
        \State $\log w_i \gets
        \log R(o_i)+\ell_i^c-\ell_i^\pi$
    \EndFor

    \State $\widehat A_i \gets
    \dfrac{w_i-\operatorname{mean}_j(w_j)}
    {\operatorname{std}_j(w_j)}$
    \State Stop gradients through $\widehat A_i$

    \For{$k=1,\ldots,K$}
        \State $\rho_{i,t}(\theta)\gets
        \dfrac{\pi_\theta(a_{i,t}\mid s_{i,t})}
        {\pi_{\mathrm{old}}(a_{i,t}\mid s_{i,t})}$

        \State $\displaystyle
        \mathcal H_\pi(\theta)\gets
        \frac{1}{G}\sum_{i=1}^{G}\frac{1}{T_i}
        \sum_{t=0}^{T_i-1}
        \mathcal H\!\left(\pi_\theta(\cdot\mid s_{i,t})\right)$

        \State $\displaystyle
        \mathcal L_\pi(\theta)\gets
        -\frac{1}{G}\sum_{i=1}^{G}\frac{1}{T_i}
        \sum_{t=0}^{T_i-1}
        \min\!\left(
            \rho_{i,t}(\theta)\widehat A_i,\,
            \operatorname{clip}
            (\rho_{i,t}(\theta),1-\epsilon,1+\epsilon)\widehat A_i
        \right)
        -\beta\,\mathcal H_\pi(\theta)$

        \State Take one optimizer step on $\theta$
    \EndFor

    \For{$k_c=1,\ldots,K_c$}
        \State $\displaystyle
        \mathcal L_c(\phi)\gets
        -\frac{1}{G}\sum_{i=1}^{G}
        \sum_{t=1}^{T_i}
        \log c_\phi(s_{i,t-1},a_{i,t-1}\mid s_{i,t})$
        \State Take one optimizer step on $\phi$
    \EndFor
\EndFor
\end{algorithmic}
\end{algorithm}

\section{Further connections to GFlowNets and symmetry}
\label{app:related}

\paragraph{Trajectory balance.}
The trajectory-conditional model $c_\phi$ represents a distribution over
construction paths, as a GFlowNet backward policy does through transitions
from a terminal object to the initial state. In trajectory
balance (TB), the forward and backward probabilities satisfy
\citep{malkin2022tb}
\[
Z P_F(\tau)=R(o)P_B(\tau\mid o).
\]
At exact balance, $P_F(\tau)/P_B(\tau\mid o)=R(o)/Z=P_F(o)$.
MIPS uses the same probability-ratio identity, but to estimate the \emph{current}
policy's marginal terminal probability for IPS, rather than to enforce this target-reward
constraint. Its trajectory-conditional model tracks the policy's conditional paths
through likelihood fitting; the policy receives the resulting IPS correction through the GRPO surrogate. 

\paragraph{Policy-gradient GFlowNets.}
GFlowNets can also be trained through policy gradients and
entropy-regularized control
\citep{niu2024gfnpg,tiapkin2024entropygfn,deleu2024multipath}.
\citet{zykovamyzina2026ppo} derive Entropic PPO for GFlowNet sampling, using
log-backward-probability rewards, entropy regularization, and optionally
likelihood-based backward-policy learning. MIPS instead augments conventional
reward-based RL: a trajectory-conditional model estimates marginal outcome
probabilities for IPS, whose scaled rewards define group-relative advantages.

\paragraph{Multiplicity and symmetry.}
\citet{kim2025symmetry} address biases from graph symmetries and equivalent
actions through symmetry-aware reward scaling. MIPS instead corrects
construction-path multiplicity when estimating outcome probabilities from
trajectory likelihoods. For outcomes defined up to symmetry, the policy and
conditional model must use the same equivalence classes; symmetry and MIPS
corrections thus address related but distinct biases.

\section{Experimental details}
\label{app:details}

\subsection{Hypergrid scalability}
\label{app:hypergrid-sweep}

\begin{table}[p]
\centering
\caption{
\textbf{Jensen--Shannon divergence on the complete Hypergrid sweep ($\downarrow$).}
Error between the exact learned terminal distribution and the
reward-distribution target.
Each entry is the mean $\pm$ standard deviation across repeated runs.
}
\label{tab:hypergrid-jsd}
\setlength{\tabcolsep}{3pt}
\resizebox{\textwidth}{!}{%
\begin{tabular}{rlrrrrrrrr}
\toprule
$H$ & Method & $G=16$ & $G=32$ & $G=64$ & $G=128$ & $G=256$ & $G=512$ & $G=1024$ & $G=2048$ \\
\midrule
4 & GRPO & $0.4535\pm0.0005$ & $0.4539\pm0.0000$ & $0.3981\pm0.1115$ & $0.3983\pm0.1113$ & $0.2985\pm0.1059$ & $0.2985\pm0.1060$ & $0.2457\pm0.0223$ & $0.2429\pm0.0235$ \\
 & IPS-GRPO & $0.1277\pm0.0006$ & $0.0420\pm0.0018$ & $0.0019\pm0.0005$ & $\boldsymbol{0.0003\pm0.0003}$ & $\boldsymbol{0.0001\pm0.0000}$ & $\boldsymbol{0.0001\pm0.0000}$ & $\boldsymbol{0.0001\pm0.0000}$ & $\boldsymbol{0.0000\pm0.0000}$ \\
 & MIPS-GRPO & $\boldsymbol{0.0014\pm0.0005}$ & $\boldsymbol{0.0006\pm0.0004}$ & $\boldsymbol{0.0004\pm0.0002}$ & $0.0006\pm0.0004$ & $0.0002\pm0.0001$ & $\boldsymbol{0.0001\pm0.0000}$ & $\boldsymbol{0.0001\pm0.0000}$ & $0.0001\pm0.0000$ \\
\midrule
8 & GRPO & $0.5062\pm0.0002$ & $0.5047\pm0.0033$ & $0.5066\pm0.0002$ & $0.5067\pm0.0002$ & $0.5068\pm0.0001$ & $0.5068\pm0.0001$ & $0.5066\pm0.0004$ & $0.4525\pm0.0625$ \\
 & IPS-GRPO & $0.2430\pm0.1222$ & $0.0999\pm0.0046$ & $0.0781\pm0.0010$ & $0.0750\pm0.0003$ & $0.0173\pm0.0027$ & $\boldsymbol{0.0005\pm0.0002}$ & $\boldsymbol{0.0001\pm0.0001}$ & $\boldsymbol{0.0001\pm0.0001}$ \\
 & MIPS-GRPO & $\boldsymbol{0.0042\pm0.0021}$ & $\boldsymbol{0.0016\pm0.0006}$ & $\boldsymbol{0.0011\pm0.0004}$ & $\boldsymbol{0.0008\pm0.0004}$ & $\boldsymbol{0.0009\pm0.0003}$ & $0.0006\pm0.0005$ & $0.0003\pm0.0001$ & $0.0002\pm0.0001$ \\
\midrule
16 & GRPO & $0.6069\pm0.0011$ & $0.6081\pm0.0003$ & $0.6086\pm0.0002$ & $0.6088\pm0.0001$ & $0.6089\pm0.0000$ & $0.6089\pm0.0000$ & $0.6089\pm0.0000$ & $0.6090\pm0.0000$ \\
 & IPS-GRPO & $0.4514\pm0.0007$ & $0.4386\pm0.0016$ & $0.4401\pm0.0002$ & $0.1091\pm0.0010$ & $0.1058\pm0.0010$ & $0.0689\pm0.0011$ & $0.0035\pm0.0002$ & $\boldsymbol{0.0003\pm0.0001}$ \\
 & MIPS-GRPO & $\boldsymbol{0.0046\pm0.0006}$ & $\boldsymbol{0.0030\pm0.0009}$ & $\boldsymbol{0.0030\pm0.0015}$ & $\boldsymbol{0.0011\pm0.0003}$ & $\boldsymbol{0.0009\pm0.0003}$ & $\boldsymbol{0.0006\pm0.0004}$ & $\boldsymbol{0.0008\pm0.0003}$ & $\boldsymbol{0.0003\pm0.0002}$ \\
\midrule
32 & GRPO & $0.5541\pm0.0039$ & $0.5557\pm0.0053$ & $0.5563\pm0.0020$ & $0.5577\pm0.0039$ & $0.5573\pm0.0016$ & $0.5572\pm0.0003$ & $0.5578\pm0.0004$ & $0.5576\pm0.0002$ \\
 & IPS-GRPO & $0.5306\pm0.0018$ & $0.5303\pm0.0009$ & $0.4540\pm0.0017$ & $0.4356\pm0.0007$ & $0.4317\pm0.0009$ & $0.1003\pm0.0011$ & $0.0914\pm0.0008$ & $0.0800\pm0.0006$ \\
 & MIPS-GRPO & $\boldsymbol{0.0064\pm0.0030}$ & $\boldsymbol{0.0043\pm0.0014}$ & $\boldsymbol{0.0033\pm0.0006}$ & $\boldsymbol{0.0018\pm0.0009}$ & $\boldsymbol{0.0014\pm0.0004}$ & $\boldsymbol{0.0010\pm0.0002}$ & $\boldsymbol{0.0011\pm0.0009}$ & $\boldsymbol{0.0005\pm0.0001}$ \\
\midrule
64 & GRPO & $0.5725\pm0.0078$ & $0.5793\pm0.0099$ & $0.5734\pm0.0046$ & $0.5753\pm0.0042$ & $0.5733\pm0.0026$ & $0.5711\pm0.0033$ & $0.5713\pm0.0025$ & $0.5747\pm0.0040$ \\
 & IPS-GRPO & $0.5347\pm0.0033$ & $0.5370\pm0.0019$ & $0.5369\pm0.0021$ & $0.5265\pm0.0014$ & $0.4647\pm0.0017$ & $0.4356\pm0.0003$ & $0.4330\pm0.0003$ & $0.2039\pm0.0939$ \\
 & MIPS & $\boldsymbol{0.0097\pm0.0009}$ & $\boldsymbol{0.0074\pm0.0018}$ & $\boldsymbol{0.0050\pm0.0011}$ & $\boldsymbol{0.0036\pm0.0008}$ & $\boldsymbol{0.0022\pm0.0007}$ & $\boldsymbol{0.0016\pm0.0003}$ & $\boldsymbol{0.0009\pm0.0001}$ & $\boldsymbol{0.0007\pm0.0001}$ \\
\midrule
128 & GRPO & $0.6712\pm0.0006$ & $0.6715\pm0.0001$ & $0.6211\pm0.1004$ & $0.6080\pm0.0742$ & $0.6084\pm0.0030$ & $0.6026\pm0.0070$ & $0.5991\pm0.0034$ & $0.5981\pm0.0041$ \\
 & IPS-GRPO & $0.5292\pm0.0025$ & $0.5312\pm0.0029$ & $0.5291\pm0.0009$ & $0.5247\pm0.0018$ & $0.5237\pm0.0011$ & $0.5200\pm0.0017$ & $0.4809\pm0.0003$ & $0.4344\pm0.0017$ \\
 & MIPS-GRPO & $\boldsymbol{0.0185\pm0.0019}$ & $\boldsymbol{0.0108\pm0.0016}$ & $\boldsymbol{0.0066\pm0.0008}$ & $\boldsymbol{0.0040\pm0.0003}$ & $\boldsymbol{0.0032\pm0.0006}$ & $\boldsymbol{0.0021\pm0.0005}$ & $\boldsymbol{0.0015\pm0.0003}$ & $\boldsymbol{0.0010\pm0.0001}$ \\
\midrule
256 & GRPO & $0.6849\pm0.0000$ & $0.6849\pm0.0000$ & $0.6849\pm0.0000$ & $0.6849\pm0.0000$ & $0.6849\pm0.0000$ & $0.6849\pm0.0000$ & $0.6849\pm0.0000$ & $0.6849\pm0.0000$ \\
 & IPS-GRPO & $0.5464\pm0.0029$ & $0.5515\pm0.0026$ & $0.5459\pm0.0032$ & $0.5437\pm0.0009$ & $0.5426\pm0.0016$ & $0.5371\pm0.0009$ & $0.5312\pm0.0027$ & $0.5261\pm0.0028$ \\
 & MIPS-GRPO & $\boldsymbol{0.0270\pm0.0026}$ & $\boldsymbol{0.0175\pm0.0025}$ & $\boldsymbol{0.0102\pm0.0017}$ & $\boldsymbol{0.0073\pm0.0005}$ & $\boldsymbol{0.0050\pm0.0010}$ & $\boldsymbol{0.0028\pm0.0002}$ & $\boldsymbol{0.0024\pm0.0005}$ & $\boldsymbol{0.0014\pm0.0002}$ \\
\midrule
512 & GRPO & $0.6902\pm0.0000$ & $0.6902\pm0.0000$ & $0.6902\pm0.0000$ & $0.6902\pm0.0000$ & $0.6902\pm0.0000$ & $0.6902\pm0.0000$ & $0.6902\pm0.0000$ & $0.6902\pm0.0000$ \\
 & IPS-GRPO & $0.5437\pm0.0029$ & $0.5650\pm0.0022$ & $0.5721\pm0.0035$ & $0.5664\pm0.0014$ & $0.5605\pm0.0006$ & $0.5585\pm0.0020$ & $0.5537\pm0.0003$ & $0.5402\pm0.0008$ \\
 & MIPS-GRPO & $\boldsymbol{0.0428\pm0.0039}$ & $\boldsymbol{0.0250\pm0.0027}$ & $\boldsymbol{0.0160\pm0.0012}$ & $\boldsymbol{0.0106\pm0.0006}$ & $\boldsymbol{0.0071\pm0.0003}$ & $\boldsymbol{0.0045\pm0.0004}$ & $\boldsymbol{0.0032\pm0.0002}$ & $\boldsymbol{0.0025\pm0.0003}$ \\
\bottomrule
\end{tabular}%
}
\end{table}

\begin{table}[p]
\centering
\caption{
\textbf{Total variation distance on the complete Hypergrid sweep ($\downarrow$).}
Error between the exact learned terminal distribution and the
reward-distribution target.
Each entry is the mean $\pm$ standard deviation across repeated runs.
}
\label{tab:hypergrid-tv}
\setlength{\tabcolsep}{3pt}
\resizebox{\textwidth}{!}{%
\begin{tabular}{rlrrrrrrrr}
\toprule
$H$ & Method & $G=16$ & $G=32$ & $G=64$ & $G=128$ & $G=256$ & $G=512$ & $G=1024$ & $G=2048$ \\
\midrule
4 & GRPO & $0.8332\pm0.0002$ & $0.8333\pm0.0000$ & $0.7708\pm0.1251$ & $0.7709\pm0.1248$ & $0.6562\pm0.1197$ & $0.6561\pm0.1198$ & $0.5982\pm0.0198$ & $0.5937\pm0.0208$ \\
 & IPS-GRPO & $0.3314\pm0.0002$ & $0.2151\pm0.0037$ & $0.0508\pm0.0084$ & $\boldsymbol{0.0193\pm0.0094}$ & $\boldsymbol{0.0111\pm0.0007}$ & $\boldsymbol{0.0085\pm0.0007}$ & $\boldsymbol{0.0084\pm0.0008}$ & $\boldsymbol{0.0051\pm0.0019}$ \\
 & MIPS-GRPO & $\boldsymbol{0.0417\pm0.0070}$ & $\boldsymbol{0.0260\pm0.0119}$ & $\boldsymbol{0.0247\pm0.0086}$ & $0.0252\pm0.0111$ & $0.0153\pm0.0054$ & $0.0126\pm0.0007$ & $0.0112\pm0.0029$ & $0.0137\pm0.0014$ \\
\midrule
8 & GRPO & $0.8837\pm0.0001$ & $0.8831\pm0.0014$ & $0.8839\pm0.0000$ & $0.8839\pm0.0000$ & $0.8839\pm0.0000$ & $0.8839\pm0.0000$ & $0.8838\pm0.0001$ & $0.8259\pm0.0670$ \\
 & IPS-GRPO & $0.5626\pm0.1608$ & $0.2948\pm0.0110$ & $0.2163\pm0.0032$ & $0.2116\pm0.0001$ & $0.1166\pm0.0064$ & $\boldsymbol{0.0236\pm0.0042}$ & $\boldsymbol{0.0119\pm0.0041}$ & $\boldsymbol{0.0120\pm0.0033}$ \\
 & MIPS-GRPO & $\boldsymbol{0.0692\pm0.0224}$ & $\boldsymbol{0.0430\pm0.0082}$ & $\boldsymbol{0.0360\pm0.0099}$ & $\boldsymbol{0.0291\pm0.0102}$ & $\boldsymbol{0.0329\pm0.0064}$ & $0.0243\pm0.0113$ & $0.0170\pm0.0020$ & $0.0175\pm0.0038$ \\
\midrule
16 & GRPO & $0.9597\pm0.0005$ & $0.9601\pm0.0001$ & $0.9602\pm0.0001$ & $0.9603\pm0.0000$ & $0.9603\pm0.0000$ & $0.9604\pm0.0000$ & $0.9604\pm0.0000$ & $0.9604\pm0.0000$ \\
 & IPS-GRPO & $0.8293\pm0.0006$ & $0.8186\pm0.0020$ & $0.8207\pm0.0001$ & $0.2922\pm0.0025$ & $0.2885\pm0.0006$ & $0.2492\pm0.0018$ & $0.0629\pm0.0024$ & $0.0204\pm0.0028$ \\
 & MIPS-GRPO & $\boldsymbol{0.0755\pm0.0077}$ & $\boldsymbol{0.0616\pm0.0112}$ & $\boldsymbol{0.0617\pm0.0173}$ & $\boldsymbol{0.0363\pm0.0040}$ & $\boldsymbol{0.0333\pm0.0062}$ & $\boldsymbol{0.0268\pm0.0087}$ & $\boldsymbol{0.0310\pm0.0064}$ & $\boldsymbol{0.0201\pm0.0044}$ \\
\midrule
32 & GRPO & $0.9179\pm0.0059$ & $0.9227\pm0.0039$ & $0.9203\pm0.0018$ & $0.9215\pm0.0006$ & $0.9222\pm0.0001$ & $0.9226\pm0.0000$ & $0.9226\pm0.0000$ & $0.9226\pm0.0000$ \\
 & IPS-GRPO & $0.9009\pm0.0020$ & $0.9007\pm0.0006$ & $0.8183\pm0.0016$ & $0.8137\pm0.0006$ & $0.8114\pm0.0011$ & $0.2845\pm0.0045$ & $0.2526\pm0.0008$ & $0.2429\pm0.0008$ \\
 & MIPS-GRPO & $\boldsymbol{0.0863\pm0.0215}$ & $\boldsymbol{0.0727\pm0.0102}$ & $\boldsymbol{0.0633\pm0.0065}$ & $\boldsymbol{0.0461\pm0.0134}$ & $\boldsymbol{0.0407\pm0.0069}$ & $\boldsymbol{0.0342\pm0.0049}$ & $\boldsymbol{0.0350\pm0.0145}$ & $\boldsymbol{0.0246\pm0.0031}$ \\
\midrule
64 & GRPO & $0.9246\pm0.0043$ & $0.9300\pm0.0066$ & $0.9284\pm0.0035$ & $0.9282\pm0.0030$ & $0.9266\pm0.0018$ & $0.9260\pm0.0027$ & $0.9264\pm0.0012$ & $0.9284\pm0.0034$ \\
 & IPS-GRPO & $0.9002\pm0.0030$ & $0.9032\pm0.0028$ & $0.9044\pm0.0026$ & $0.8959\pm0.0012$ & $0.8308\pm0.0024$ & $0.8126\pm0.0002$ & $0.8122\pm0.0002$ & $0.4926\pm0.1538$ \\
 & MIPS-GRPO & $\boldsymbol{0.1076\pm0.0044}$ & $\boldsymbol{0.0957\pm0.0142}$ & $\boldsymbol{0.0789\pm0.0084}$ & $\boldsymbol{0.0674\pm0.0075}$ & $\boldsymbol{0.0518\pm0.0077}$ & $\boldsymbol{0.0439\pm0.0033}$ & $\boldsymbol{0.0328\pm0.0029}$ & $\boldsymbol{0.0291\pm0.0022}$ \\
\midrule
128 & GRPO & $0.9868\pm0.0004$ & $0.9870\pm0.0001$ & $0.9467\pm0.0803$ & $0.9401\pm0.0559$ & $0.9465\pm0.0019$ & $0.9429\pm0.0032$ & $0.9401\pm0.0011$ & $0.9386\pm0.0028$ \\
 & IPS-GRPO & $0.8905\pm0.0031$ & $0.8949\pm0.0029$ & $0.8942\pm0.0017$ & $0.8927\pm0.0026$ & $0.8937\pm0.0017$ & $0.8915\pm0.0011$ & $0.8501\pm0.0004$ & $0.8089\pm0.0022$ \\
 & MIPS-GRPO & $\boldsymbol{0.1515\pm0.0107}$ & $\boldsymbol{0.1137\pm0.0091}$ & $\boldsymbol{0.0872\pm0.0056}$ & $\boldsymbol{0.0679\pm0.0026}$ & $\boldsymbol{0.0625\pm0.0067}$ & $\boldsymbol{0.0492\pm0.0070}$ & $\boldsymbol{0.0422\pm0.0047}$ & $\boldsymbol{0.0338\pm0.0018}$ \\
\midrule
256 & GRPO & $0.9955\pm0.0000$ & $0.9955\pm0.0000$ & $0.9955\pm0.0000$ & $0.9955\pm0.0000$ & $0.9955\pm0.0000$ & $0.9955\pm0.0000$ & $0.9955\pm0.0000$ & $0.9955\pm0.0000$ \\
 & IPS-GRPO & $0.8975\pm0.0017$ & $0.9067\pm0.0028$ & $0.9068\pm0.0032$ & $0.9065\pm0.0014$ & $0.9071\pm0.0018$ & $0.9039\pm0.0013$ & $0.8997\pm0.0032$ & $0.8957\pm0.0019$ \\
 & MIPS-GRPO & $\boldsymbol{0.1814\pm0.0106}$ & $\boldsymbol{0.1435\pm0.0106}$ & $\boldsymbol{0.1100\pm0.0095}$ & $\boldsymbol{0.0945\pm0.0042}$ & $\boldsymbol{0.0766\pm0.0085}$ & $\boldsymbol{0.0568\pm0.0028}$ & $\boldsymbol{0.0526\pm0.0060}$ & $\boldsymbol{0.0409\pm0.0024}$ \\
\midrule
512 & GRPO & $0.9985\pm0.0000$ & $0.9985\pm0.0000$ & $0.9985\pm0.0000$ & $0.9985\pm0.0000$ & $0.9985\pm0.0000$ & $0.9985\pm0.0000$ & $0.9985\pm0.0000$ & $0.9985\pm0.0000$ \\
 & IPS-GRPO & $0.8967\pm0.0015$ & $0.9108\pm0.0018$ & $0.9194\pm0.0022$ & $0.9186\pm0.0012$ & $0.9165\pm0.0003$ & $0.9157\pm0.0014$ & $0.9130\pm0.0002$ & $0.9054\pm0.0014$ \\
 & MIPS-GRPO & $\boldsymbol{0.2323\pm0.0098}$ & $\boldsymbol{0.1735\pm0.0089}$ & $\boldsymbol{0.1390\pm0.0061}$ & $\boldsymbol{0.1129\pm0.0041}$ & $\boldsymbol{0.0917\pm0.0021}$ & $\boldsymbol{0.0720\pm0.0039}$ & $\boldsymbol{0.0604\pm0.0027}$ & $\boldsymbol{0.0541\pm0.0034}$ \\
\bottomrule
\end{tabular}%
}
\end{table}

\begin{table}[t]
\centering
\caption{Hypergrid reproducibility settings.
\(H\) and \(G\) are swept; every other value is fixed.
Seeds are \(1,2,3\).}
\label{tab:mips-hypergrid}
\begin{tabular}{ll}
\toprule
Setting & Value \\
\midrule
Grid & \(D=2\), \(H\in\{4,8,16,32,64,128,256,512\}\) \\
Coordinates & \(\mathrm{linspace}(-1,1,H)\) \\
Actions & Increment one coordinate, or stop \\
Termination & Stop, or all coordinates at \(H-1\) \\
Policy & MLP, \(HD \to 128 \to 128 \to D+1\), ReLU \\
Optimizer & Adam, step size \(10^{-3}\), \((\beta_1,\beta_2)=(0.9,0.999)\) \\
Updates & \(10{,}000\), one full-batch step each \\
Batch size \(G\) & \(\{16,32,64,128,256,512,1024,2048\}\) \\
Entropy coefficient & \(0.1\) \\
\bottomrule
\end{tabular}
\end{table}

\FloatBarrier

\subsection{Molecular synthesis}
\label{app:molecule}

\begin{table}[htbp]
\centering
\small
\caption{\textbf{Hyperparameters for molecular synthesis.} GRPO, IPS-GRPO,
and MIPS-GRPO share the same policy architecture and training settings;
MIPS-GRPO additionally trains a separate conditional model $c_\phi$. RGFN
uses its published configuration.}
\label{tab:molecule-hparams}
\begin{tabular}{ll}
\toprule
Hyperparameter & Value \\
\midrule
\multicolumn{2}{l}{\textit{Shared}} \\
Policy network & Graph transformer, $5$ layers, $4$ heads, hidden dim $64$ \\
Graph features & Node dim $71$, edge dim $4$ \\
Output heads & Phase-specific MLPs with GELU \\
Maximum reactions & $4$ \\
Reward & $\exp(\beta \cdot \text{proxy})$, $\beta = 8$ \\
Optimizer & Adam \\
Training iterations & $4{,}000$ \\
Group size $G$ & $100$ \\
Replay buffer & None \\
Exploration & None (on-policy) \\
Policy-ratio clip $\epsilon$ & $0.2$ \\
Gradient clip (policy) & $10.0$ \\
Learning rate (GRPO, IPS-GRPO) & $1 \times 10^{-3}$ \\
\midrule
\multicolumn{2}{l}{\textit{MIPS-GRPO}} \\

Conditional model $c_\phi$ & Separate network, same architecture \\
Learning rate ($c_\phi$) & $1 \times 10^{-3}$ \\
Fitting epochs per update ($c_\phi$) & $4$ \\
Gradient clip ($c_\phi$) & $1.0$ \\
\bottomrule
\end{tabular}
\end{table}

\paragraph{Evaluation details.}

In addition to the metrics in the main text, we report four quantities. Mode candidates are unique molecules with a proxy score above $7$ before similarity filtering. Thus, if a method has no mode candidates, it has not sampled any high-reward molecules. Following \citet{koziarski2024rgfn}, we also count Murcko scaffolds among molecules with proxy score above $8$. For the discovered modes, we report two standard chemistry metrics: quantitative estimate of drug-likeness (QED) and synthetic accessibility (SA). QED ranges from $0$ to $1$, with higher values indicating greater drug-likeness. SA ranges from $1$ to $10$, with lower values indicating easier synthesis. Our evaluation differs from \citet{koziarski2024rgfn} in that we count modes using samples from the final policy rather than all molecules visited during training.

\begin{table}[htbp]
\centering
\small
\caption{\textbf{Per-seed molecular synthesis results.} Statistics of
$50{,}000$ samples from each final policy (3 seeds). Unique fractions are computed over
valid samples, and scaffold counts are taken among molecules above the given
proxy score. Mode-level quality metrics are averaged over discovered modes and
are undefined when no mode is found.}
\label{tab:molecule-seeds-div}
\label{tab:molecule-seeds-qual}
\resizebox{\textwidth}{!}{%
\begin{tabular}{lcrrrrrrrrrrrr}
\toprule
& & \multicolumn{7}{c}{Diversity} & \multicolumn{4}{c}{Quality} \\
\cmidrule(lr){3-9}\cmidrule(lr){10-13}
Method & Seed & Unique & Unique (\%) & Modes & Candidates
  & Scaffolds ($>7$) & Scaffolds ($>8$) & Valid (\%)
  & Mean proxy & Mode proxy & Mode QED & Mode SA \\
\midrule
GRPO      & 0 & $1$ & $0.002$ & $0$ & $0$ & $0$ & $0$ & $100.0$ & $6.60$ & --- & --- & --- \\
          & 1 & $1$ & $0.002$ & $0$ & $0$ & $0$ & $0$ & $100.0$ & $6.88$ & --- & --- & --- \\
          & 2 & $1$ & $0.002$ & $0$ & $0$ & $0$ & $0$ & $100.0$ & $6.15$ & --- & --- & --- \\
\midrule
IPS-GRPO  & 0 & $6$ & $0.012$ & $0$ & $0$ & $0$ & $0$ & $100.0$ & $3.35$ & --- & --- & --- \\
          & 1 & $17$ & $0.034$ & $0$ & $0$ & $0$ & $0$ & $100.0$ & $5.97$ & --- & --- & --- \\
          & 2 & $27$ & $0.055$ & $0$ & $0$ & $0$ & $0$ & $100.0$ & $4.72$ & --- & --- & --- \\
\midrule
RGFN      & 0 & $35{,}711$ & $71.5$ & $19$ & $27$ & $25$ & $0$ & $99.9$ & $4.41$ & $7.12$ & $0.234$ & $3.11$ \\
          & 1 & $33{,}306$ & $66.6$ & $17$ & $36$ & $30$ & $0$ & $100.0$ & $4.40$ & $7.22$ & $0.319$ & $3.01$ \\
          & 2 & $35{,}216$ & $70.5$ & $18$ & $19$ & $19$ & $0$ & $100.0$ & $3.76$ & $7.16$ & $0.369$ & $2.95$ \\
\midrule
MIPS-GRPO & 0 & $47{,}022$ & $94.1$ & $71$ & $132$ & $128$ & $1$ & $100.0$ & $4.39$ & $7.30$ & $0.441$ & $2.87$ \\
          & 1 & $43{,}105$ & $86.2$ & $72$ & $236$ & $197$ & $1$ & $100.0$ & $4.62$ & $7.21$ & $0.289$ & $2.80$ \\
          & 2 & $47{,}255$ & $95.2$ & $68$ & $122$ & $122$ & $0$ & $100.0$ & $4.40$ & $7.15$ & $0.301$ & $2.91$ \\
\bottomrule
\end{tabular}%
}
\end{table}

\FloatBarrier

\subsection{Phylogenetic inference}
\label{app:phylo}

\paragraph{Environment details.}
We use the first $10$ taxa of the DS1 alignment ($1{,}949$ sites).
Branch lengths are chosen from $50$ discrete values, $0.001, 0.002, \ldots, 0.050$, for all methods.
 
\paragraph{Training details.}
All four methods share the forward network. GRPO, IPS-GRPO, and MIPS-GRPO
each process $4{,}096$ on-policy trajectories per iteration for $10{,}000$
iterations, i.e.\ $40.96$M trajectories without replay. PhyloGFN processes
$3{,}277$ on-policy and $819$ replayed trajectories per iteration for the
same $10{,}000$ iterations, also $40.96$M in total. MIPS-GRPO additionally
trains a separate conditional model $c_\phi$.
Table~\ref{tab:phylo-hparams-10taxa} lists all hyperparameters.

 \begin{table}[htbp]
\centering
\small
\caption{\textbf{Hyperparameters for phylogenetic inference (DS1 subset,
$10$ taxa).}}
\label{tab:phylo-hparams-10taxa}
\begin{tabular}{lp{0.58\linewidth}}
\toprule
Hyperparameter & Value \\
\midrule
\multicolumn{2}{l}{\textit{Shared (all four methods)}} \\
Policy network & Transformer, $6$ layers, $4$ heads; MLP output heads,
  hidden $256$ \\
Training iterations & $10{,}000$ \\
Exploration probability & Linear from $0.5$ to $0.001$ over $200$
  iterations \\
\midrule
\multicolumn{2}{l}{\textit{GRPO, IPS-GRPO, MIPS-GRPO}} \\
Group size $G$ & $4{,}096$ \\
Policy epochs $K$ & $1$ \\
Learning rate (policy) & $10^{-4}$ (Adam) \\
Clip parameter $\epsilon$ & $0.2$ \\
Entropy coefficient $\beta$ & $0$ \\
\midrule
\multicolumn{2}{l}{\textit{MIPS-GRPO}} \\
Conditional model $c_\phi$ & MLP, $2$ hidden layers of width $128$, over
  valid merges; input features describe the step, the merged subtrees, and
  the terminal tree \\
Conditional-model epochs $K_c$ & $4$ \\
Learning rate ($c_\phi$) & $10^{-3}$ (Adam) \\
\midrule
\multicolumn{2}{l}{\textit{PhyloGFN}} \\
Loss / backward policy & Trajectory balance / fixed uniform \\
Learning rate (model / $\log Z$) & $10^{-4}$ / $10^{-2}$ (Adam) \\
\bottomrule
\end{tabular}
\end{table}

\begin{figure}[htbp]
    \centering
    \includegraphics[width=\linewidth]{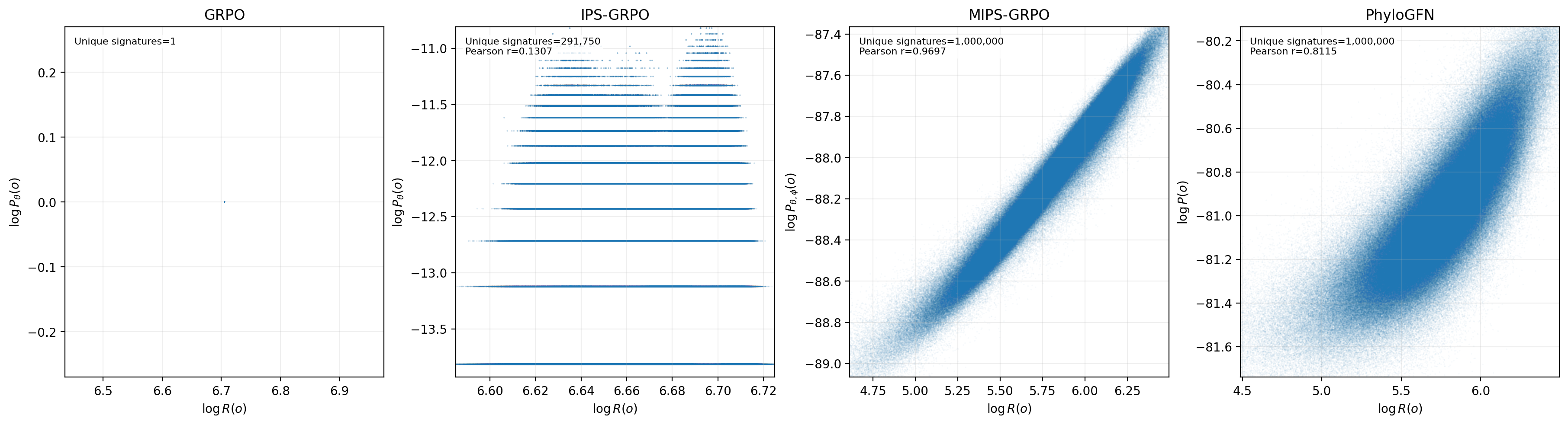}
    \caption{\textbf{Outcome log-probability versus log-reward.}
Each panel shows $10^6$ trees sampled from the final policy.
GRPO places all probability on a single tree ($\log P_\theta(o)=0$), and IPS-GRPO's probabilities barely vary with reward.
MIPS-GRPO and PhyloGFN cover a broad set of trees, with MIPS-GRPO's estimated probability $P_{\theta,\phi}(o)$ following the reward most closely.
Axis ranges differ across panels.}
    \label{fig:phylo-sampling-calibration-10taxa}
\end{figure}
 
\begin{table}[htbp]
\centering
\small
\caption{\textbf{Per-seed phylogenetic results.} Unique counts among $10^6$
samples from each final policy.}
\label{tab:phylo-seeds-10taxa}
\resizebox{\textwidth}{!}{%
\begin{tabular}{lcrrcccc}
\toprule
Method & Seed & Unique topologies & Unique signatures & Sampling $r$
  & Marginal log-lik. & MLL & Shared-tree $r$ \\
\midrule
GRPO      & 0 & $1$ & $1$ & --- & $-4318.36$ & $-4183.27$ & --- \\
          & 1 & $1$ & $1$ & --- & $-4318.43$ & $-4182.40$ & --- \\
          & 2 & $1$ & $1$ & --- & $-4318.43$ & $-4182.61$ & --- \\
\midrule
IPS-GRPO  & 0 & $13$ & $291{,}750$ & $0.131$ & $-4293.30$ & $-4212.82$ & --- \\
          & 1 & $11$ & $248{,}245$ & $0.150$ & $-4298.43$ & $-4214.99$ & --- \\
          & 2 & $13$ & $250{,}600$ & $0.140$ & $-4291.36$ & $-4211.82$ & --- \\
\midrule
PhyloGFN  & 0 & $985{,}326$ & $1{,}000{,}000$ & $0.812$ & $-4367.72$ & $-4672.82$ & $0.942$ \\
          & 1 & $985{,}263$ & $1{,}000{,}000$ & $0.886$ & $-4363.99$ & $-4671.59$ & $0.922$ \\
          & 2 & $985{,}322$ & $1{,}000{,}000$ & $0.809$ & $-4367.77$ & $-4672.78$ & $0.943$ \\
\midrule
MIPS-GRPO & 0 & $955{,}839$ & $1{,}000{,}000$ & $0.970$ & $-4372.49$ & $-4663.80$ & $0.971$ \\
          & 1 & $955{,}918$ & $1{,}000{,}000$ & $0.968$ & $-4373.98$ & $-4664.66$ & $0.952$ \\
          & 2 & $955{,}800$ & $1{,}000{,}000$ & $0.964$ & $-4372.22$ & $-4663.60$ & $0.964$ \\
\bottomrule
\end{tabular}%
}
\end{table}

\FloatBarrier

\end{document}